\documentclass[10pt,twocolumn,letterpaper]{article}

\usepackage{cvpr}
\usepackage{times}
\usepackage{epsfig}
\usepackage{graphicx}
\usepackage{amsmath}
\usepackage{amssymb}

\usepackage{tabulary}
\usepackage{multirow}

\usepackage{color}

\usepackage{wrapfig}
\usepackage{enumitem} 
\usepackage{adjustbox}

\definecolor{citecolor}{RGB}{34,139,34}
\usepackage[pagebackref=true,breaklinks=true,letterpaper=true,colorlinks,citecolor=citecolor,bookmarks=false]{hyperref}

 \cvprfinalcopy 

\newcommand{\app}{\raise.17ex\hbox{$\scriptstyle\sim$}}

\makeatletter\renewcommand\paragraph{\@startsection{paragraph}{4}{\z@}
  {.5em \@plus1ex \@minus.2ex}{-.5em}{\normalfont\normalsize\bfseries}}\makeatother

\newlength\savewidth\newcommand\shline{\noalign{\global\savewidth\arrayrulewidth
  \global\arrayrulewidth 1pt}\hline\noalign{\global\arrayrulewidth\savewidth}}

\makeatletter\renewcommand\paragraph{\@startsection{paragraph}{4}{\z@}
  {.5em \@plus1ex \@minus.2ex}{-.5em}{\normalfont\normalsize\bfseries}}\makeatother

\def\cvprPaperID{***} 

\begin{document}

\title{\vspace{-.5em} Rethinking ImageNet Pre-training \\ \vspace{-.5em}}

\author{
 Kaiming He \quad Ross Girshick \quad Piotr Doll\'ar \vspace{3mm}\\
 Facebook AI Research (FAIR)
}

\maketitle

\begin{abstract}
We report competitive results on object detection and instance segmentation on the COCO dataset using standard models trained \textbf{from random initialization}. The results are \textbf{no worse} than their ImageNet pre-training counterparts even when using the hyper-parameters of the baseline system (Mask R-CNN) that were optimized for fine-tuning pre-trained models, with the sole exception of increasing the number of training iterations so the randomly initialized models may converge. Training from random initialization is surprisingly robust; our results hold even when: (i) using only 10\% of the training data, (ii) for deeper and wider models, and (iii) for multiple tasks and metrics. Experiments show that ImageNet pre-training speeds up convergence early in training, but does not necessarily provide regularization or improve final target task accuracy. To push the envelope we demonstrate {50.9}~AP on COCO object detection without using any external data---a result on par with the top COCO 2017 competition results that used ImageNet pre-training. These observations challenge the conventional wisdom of ImageNet pre-training for dependent tasks and we expect these discoveries will encourage people to rethink the current de facto paradigm of `pre-training and fine-tuning' in computer vision.
\end{abstract}

\vspace{-.5em}
\section{Introduction}

Deep convolutional neural networks \cite{Krizhevsky2012,LeCun1989} revolutionized computer vision arguably due to the discovery that feature representations learned on a \emph{pre-training} task can transfer useful information to target tasks \cite{Girshick2014,Donahue2014,Zeiler2014}. In recent years, a well-established paradigm has been to pre-train models using large-scale data (\eg, ImageNet \cite{Russakovsky2015}) and then to fine-tune the models on target tasks that often have less training data. Pre-training has enabled state-of-the-art results on many tasks, including object detection \cite{Girshick2014,Girshick2015,Ren2015}, image segmentation \cite{Long2015,He2017}, and action recognition \cite{Simonyan2014,Carreira2017}.

\begin{figure}[t]
\centering
\includegraphics[height=19.8em]{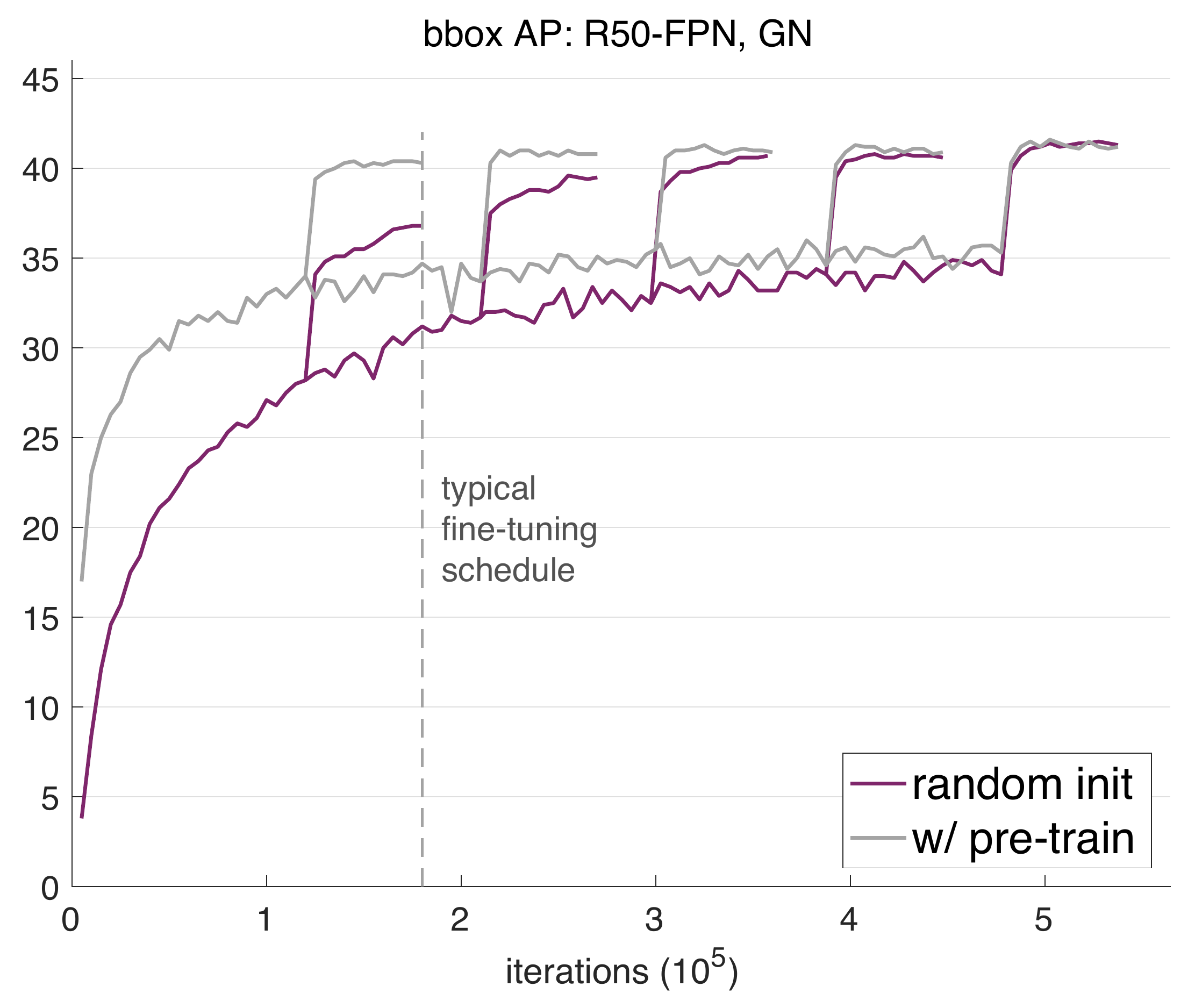}
\caption{We train Mask R-CNN \cite{He2017} with a ResNet-50 FPN \cite{Lin2017} and GroupNorm \cite{Wu2018} backbone on the COCO \texttt{train2017} set and evaluate bounding box AP on the \texttt{val2017} set, initializing the model by random weights or ImageNet pre-training. We explore different training schedules by varying the iterations at which the learning rate is reduced (where the accuracy leaps). 
The model trained from random initialization needs more iterations to converge, but converges to a solution that is \emph{no worse than the fine-tuning counterpart}. Table~\ref{tab:coco_gn_sched} shows the resulting AP numbers.}\label{fig:bbox_r50_gn}
\vspace{-1em}
\end{figure}

A path to `solving' computer vision then appears to be paved by pre-training a `universal' feature representation on ImageNet-like data at \emph{massive} scale \cite{Sun2017,Mahajan2018}. Attempts along this path have pushed the frontier to up to 3000$\times$ \cite{Mahajan2018} the size of ImageNet. However, the success of these experiments is mixed: although improvements have been observed, for object detection in particular they are small and scale poorly with the pre-training dataset size. That this path will `solve' computer vision is open to doubt.

This paper questions the paradigm of pre-training even further by exploring the opposite regime: we report that competitive object detection and instance segmentation accuracy is achievable when training on COCO \emph{from random initialization (`from scratch'), without any pre-training}. More surprisingly, we can achieve these results by \emph{using baseline systems \emph{\cite{Girshick2015,Ren2015,Lin2017,He2017}} and their hyper-parameters that were optimized for fine-tuning pre-trained models}. We find that there is no fundamental obstacle preventing us from training from scratch, if: (i) we use normalization techniques appropriately for optimization, and (ii) we train the models sufficiently long to compensate for the lack of pre-training (Figure~\ref{fig:bbox_r50_gn}).

We show that training from random initialization on COCO can be \emph{on par with} its ImageNet pre-training counterparts for a variety of baselines that cover Average Precision (AP, in percentage) from 40 to over 50. Further, we find that such comparability holds even if we train with as little as 10\% COCO training data. We also find that we can train large models from scratch---up to 4$\times$ larger than a ResNet-101 \cite{He2016}---without overfitting. Based on these experiments and others, we observe the following:

(i) ImageNet pre-training \emph{speeds up} convergence, especially early on in training, but training from random initialization can catch up after training for a duration that is roughly comparable to the total ImageNet pre-training plus fine-tuning computation---it has to learn the low-/mid-level features (such as edges, textures) that are otherwise given by pre-training. As the cost of ImageNet pre-training is often ignored when studying the target task, `controlled' comparisons with a \emph{short} training schedule can veil the true behavior of training from random initialization.

(ii) ImageNet pre-training does \emph{not} automatically give better regularization. When training with fewer images (down to 10\% of COCO), we find that new hyper-parameters must be selected for fine-tuning (from pre-training) to avoid overfitting. Then, when training from random initialization using these \emph{same} hyper-parameters, the model can match the pre-training accuracy \emph{without} any extra regularization, even with only 10\% COCO data.

(iii) ImageNet pre-training shows no benefit when the target tasks/metrics are more sensitive to spatially well-localized predictions. We observe a noticeable AP improvement for high box overlap thresholds when training from scratch; we also find that keypoint AP, which requires fine spatial localization, converges relatively faster from scratch. Intuitively, the task gap between the \emph{classification}-based, ImageNet-like pre-training and \emph{localization}-sensitive target tasks may limit the benefits of pre-training.

Given the current literature, these results are surprising and challenge our understanding of the effects of ImageNet pre-training. These observations hint that ImageNet pre-training is a historical workaround (and will likely be so for some time) for when the community does not have enough target data or computational resources to make training on the target task doable. In addition, ImageNet has been largely thought of as a `free' resource, thanks to the readily conducted annotation efforts \emph{and} wide availability of pre-trained models. But looking forward, when the community will proceed with more data and faster computation, our study suggests that collecting data and training on the \emph{target} tasks is a solution worth considering, especially when there is a significant gap between the source pre-training task and the target task. This paper provides new experimental evidence and discussions for people to rethink the ImageNet-like pre-training paradigm in computer vision.

\section{Related Work}

\paragraph{Pre-training and fine-tuning.}
The initial breakthrough of applying deep learning to object detection (\eg, R-CNN \cite{Girshick2014} and OverFeat \cite{Sermanet2014}) were achieved by fine-tuning networks that were pre-trained for ImageNet classification. Following these results, most modern object detectors and many other computer vision algorithms employ the `pre-training and fine-tuning' paradigm. 
Recent work pushes this paradigm further by pre-training on datasets that are 6$\times$ (ImageNet-5k~\cite{He2018pami}), 300$\times$ (JFT~\cite{Sun2017}), and even 3000$\times$ (Instagram~\cite{Mahajan2018}) larger than ImageNet. While this body of work demonstrates significant improvements on image classification transfer learning tasks, the improvements on object detection are relatively small (on the scale of +1.5 AP on COCO with 3000$\times$ larger pre-training data~\cite{Mahajan2018}). The marginal benefit from the kind of large-scale pre-training data used to date diminishes rapidly.

\paragraph{Detection from scratch.} Before the prevalence of the `pre-training and fine-tuning' paradigm, object detectors were trained \emph{with no pre-training} (\eg, \cite{Matan1992,Rowley1996,Szegedy2013})---a fact that is somewhat overlooked today. In fact, \emph{it should not be surprising that object detectors can be trained from scratch}.

Given the success of pre-training in the R-CNN paper \cite{Girshick2014}, later analysis \cite{Agrawal2014} found that pre-training plays an important role in detector accuracy when training data is limited, but also illustrated that \emph{training from scratch on more detection data is possible} and can achieve 90\% of the fine-tuning accuracy, foreshadowing our results.

As modern object detectors \cite{Girshick2014,He2014,Girshick2015,Ren2015,Redmon2016,Liu2016,Lin2017,He2017} evolved under the pre-training paradigm, the belief that training from scratch is non-trivial became conventional wisdom. Shen \etal \cite{Shen2017} argued for \emph{a set of new design principles} to obtain a detector that is optimized for the accuracy when trained from scratch. They designed a specialized detector driven by deeply supervised networks \cite{Lee2015} and dense connections \cite{Huang2017}. 
DetNet~\cite{Li2018} and CornerNet~\cite{Law2018} also present results when training detectors from scratch. Similar to~\cite{Shen2017}, these works \cite{Li2018,Law2018} focus on designing detection-specific architectures.
However, in \cite{Shen2017,Li2018,Law2018}  \emph{there is little evidence that these specialized architectures are required for models to be trained from scratch}. 

Unlike these papers, our focus is on understanding the role of ImageNet pre-training on \emph{unspecialized} architectures (\ie, models that were originally designed \emph{without} the consideration for training from scratch). Our work demonstrates that it is often possible to \emph{match} fine-tuning accuracy when training from scratch even \emph{without} making any architectural specializations. Our study is on the comparison between `with \vs without pre-training', under controlled settings in which the architectures are not tailored.

\section{Methodology}\label{sec:method}

Our goal is to \emph{ablate} the role of ImageNet pre-training via controlled experiments that can be done \emph{without} ImageNet pre-training. Given this goal, architectural improvements are \emph{not} our purpose; actually, to better understand what impact ImageNet pre-training can make, it is desired to enable typical architectures to be trained from scratch under \emph{minimal} modifications. We describe the only two modifications that we find to be necessary, related to model normalization and training length, discussed next.

\subsection{Normalization}

Image classifier training requires \emph{normalization} to help optimization.
Successful forms of normalization include normalized parameter initialization \cite{Glorot2010,He2015} and activation normalization layers \cite{Ioffe2015,Ba2016,Ulyanov2016,Wu2018}.
When training object detectors from scratch, they face issues similar to training image classifiers from scratch \cite{Glorot2010,He2015,Ioffe2015}.
Overlooking the role of normalization can give the misperception that detectors are hard to train from scratch.

Batch Normalization (BN) \cite{Ioffe2015}, the popular normalization method used to train modern networks, partially makes training detectors from scratch difficult.
Object detectors are typically trained with high resolution inputs, unlike image classifiers. This reduces batch sizes as constrained by memory, and small batch sizes severely degrade the accuracy of BN \cite{Ioffe2017,Peng2018,Wu2018}. This issue can be circumvented if pre-training is used, because fine-tuning can adopt the pre-training batch statistics as fixed parameters \cite{He2016}; however, freezing BN is invalid when training from scratch.

We investigate two normalization strategies in recent works that help relieve the small batch issue:
\vspace{-.4em}
\begin{enumerate}[label=(\roman*)]
\item \emph{Group Normalization} (\textbf{GN}) \cite{Wu2018}: as a recently proposed alternative to BN, GN performs computation that is independent of the batch dimension. GN's accuracy is insensitive to batch sizes \cite{Wu2018}. 
\vspace{-.2em}
\item \emph{Synchronized Batch Normalization} (\textbf{SyncBN}) \cite{Peng2018,Liu2018}: this is an implementation of BN \cite{Ioffe2015} with batch statistics computed across multiple devices (GPUs). This increases the effective batch size for BN when using many GPUs, which avoids small batches.
\end{enumerate}
\vspace{-.4em}
\noindent Our experiments show that both GN and SyncBN can enable detection models to train from scratch.

We also report that using appropriately normalized initialization \cite{He2015}, we can train object detectors with VGG nets~\cite{Simonyan2015} from random initialization without BN or GN.

\begin{figure}[t]
\centering
\includegraphics[width=1.\linewidth]{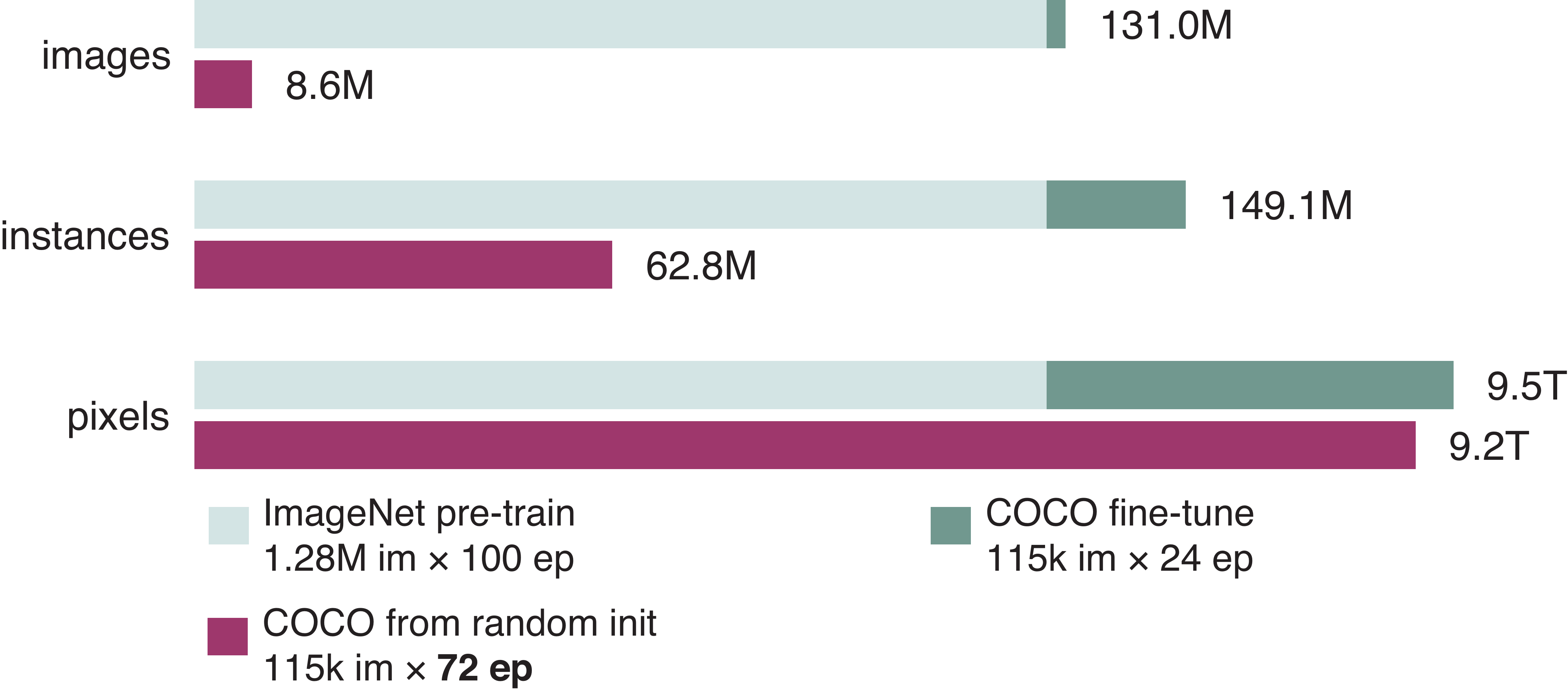}
\caption{Total numbers of images, instances, and pixels seen during all training iterations, for pre-training + fine-tuning (green bars) \vs from random initialization (purple bars). We consider that pre-training takes 100 epochs in ImageNet, and fine-tuning adopts the 2$\times$ schedule ($\app$24 epochs over COCO) and random initialization adopts the 6$\times$ schedule ($\app$72 epochs over COCO). We count instances in ImageNet as 1 per image (\vs \app7 in COCO), and pixels in ImageNet as 224$\times$224 and COCO as 800$\times$1333.
}\label{fig:stats}
\vspace{-.8em}
\end{figure}

\subsection{Convergence}

It is unrealistic and unfair to expect models trained from random initialization to converge similarly fast as those initialized from ImageNet pre-training. Overlooking this fact one can draw incomplete or incorrect conclusions about the true \emph{capability} of models that are trained from scratch.

Typical ImageNet pre-training involves over one million images iterated for one hundred epochs.
In addition to any semantic information learned from this large-scale data, the pre-training model has also learned low-level features (\eg, edges, textures) that do not need to be re-learned during fine-tuning.\footnote{In fact, it is common practice \cite{Girshick2015,Ren2015} to freeze the convolutional filters in the first few layers when fine-tuning.}
On the other hand, when training from scratch the model has to learn low- and high-level semantics, so more iterations may be necessary for it to converge well.

With this motivation, we argue that models trained from scratch must be \textbf{trained for longer} than typical fine-tuning schedules. Actually, this is a \emph{fairer} comparison in term of the number of training samples provided. We consider three rough definitions of `samples'---the number of images, instances, and pixels that have been seen during all training iterations (\eg, one image for 100 epochs is counted as 100 image-level samples). We plot the comparisons on the numbers of samples in Figure~\ref{fig:stats}.

Figure~\ref{fig:stats} shows a from-scratch case trained for 3 times more iterations than its fine-tuning counterpart on COCO. Despite using more iterations on COCO, if counting image-level samples, the from-scratch case still sees considerably \emph{fewer} samples than its fine-tuning counterpart---the 1.28 million ImageNet images for 100 epochs dominate. Actually, the sample numbers only get closer if we count \emph{pixel-level} samples (Figure~\ref{fig:stats}, bottom)---a consequence of object detectors using higher-resolution images. Our experiments show that under the schedules in Figure~\ref{fig:stats}, the from-scratch detectors can catch up with their fine-tuning counterparts. 
This suggests that a sufficiently large number of total samples (arguably in terms of pixels) are required for the models trained from random initialization to converge well.

\section{Experimental Settings}\label{sec:settings}

\newcommand{\Detectron}{\textsf{\small Detectron}\xspace}

We pursue \emph{minimal} changes made to baseline systems for pinpointing the keys to enabling training from scratch. Overall, our baselines and hyper-parameters follow Mask R-CNN \cite{He2017} in the publicly available code of \Detectron \cite{Detectron2018}, except we use normalization and vary the number of training iterations. The implementation is as follows.

\paragraph{Architecture.}
We investigate Mask R-CNN \cite{He2017} with ResNet \cite{He2016} or ResNeXt \cite{Xie2017} plus Feature Pyramid Network (FPN) \cite{Lin2017} backbones. We adopt the end-to-end fashion \cite{Ren2017} of training Region Proposal Networks (RPN) jointly with Mask R-CNN.
GN/SyncBN is used to replace all `frozen BN' (channel-wise affine) layers. For fair comparisons, in this paper the \emph{fine-tuned} models (with pre-training) are also tuned with GN or SyncBN, rather than freezing them. They have higher accuracy than the frozen ones \cite{Peng2018,Liu2018,Wu2018}.

\paragraph{Learning rate scheduling.}
Original Mask R-CNN models in \Detectron \cite{Detectron2018} were fine-tuned with 90k iterations (namely, `1$\times$ schedule') or 180k iterations (`2$\times$ schedule'). For models in this paper, we investigate longer training and we use similar terminology, \eg, a so-called `6$\times$ schedule' has 540k iterations. Following the strategy in the 2$\times$ schedule, we always reduce the learning rate by 10$\times$ in the last 60k and last 20k iterations respectively, no matter how many total iterations (\ie, the reduced learning rates are always run for the same number of  iterations). We find that training longer for the first (large) learning rate is useful, but training for longer on small learning rates often leads to overfitting.

\paragraph{Hyper-parameters.} All other hyper-parameters follow those in \Detectron \cite{Detectron2018}. Specially, the initial learning rate is 0.02 (with a linear warm-up \cite{Goyal2017}). The weight decay is 0.0001 and momentum is 0.9. All models are trained in 8 GPUs using synchronized SGD, with a mini-batch size of 2 images per GPU.

By default Mask R-CNN in \Detectron uses \emph{no data augmentation} for testing, and only horizontal flipping augmentation for training. We use the same settings. Also, unless noted, the image scale is 800 pixels for the shorter side.

\begin{figure}[t]
\centering
\includegraphics[height=19.8em]{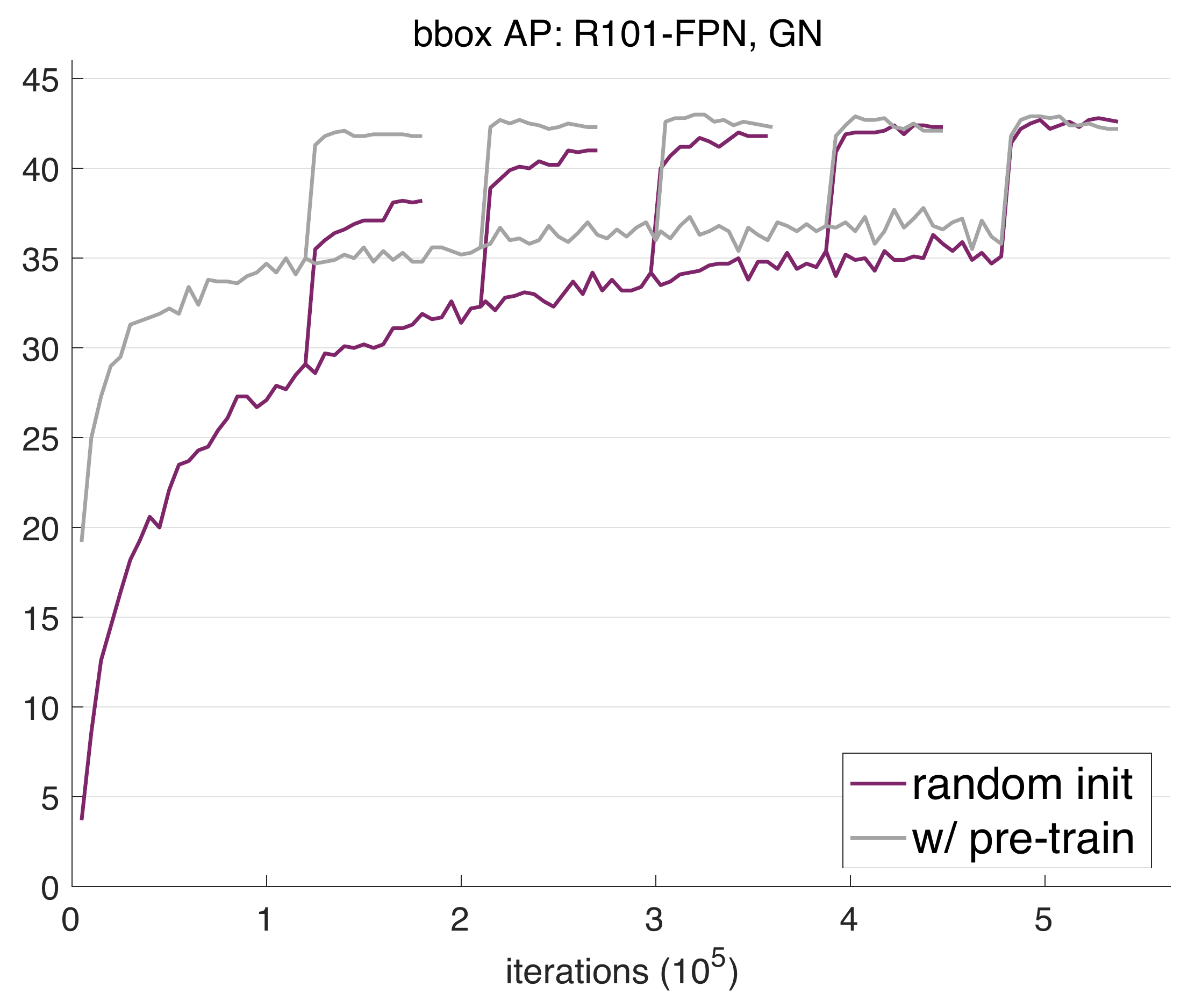}
\caption{Learning curves of AP$^\text{bbox}$ on COCO \texttt{val2017} using Mask R-CNN with \textbf{R101}-FPN and GN. Table~\ref{tab:coco_gn_sched} shows the resulting AP numbers.}\label{fig:bbox_r101_gn}
\end{figure}

\begin{figure}[t]
\centering
\includegraphics[height=19.8em]{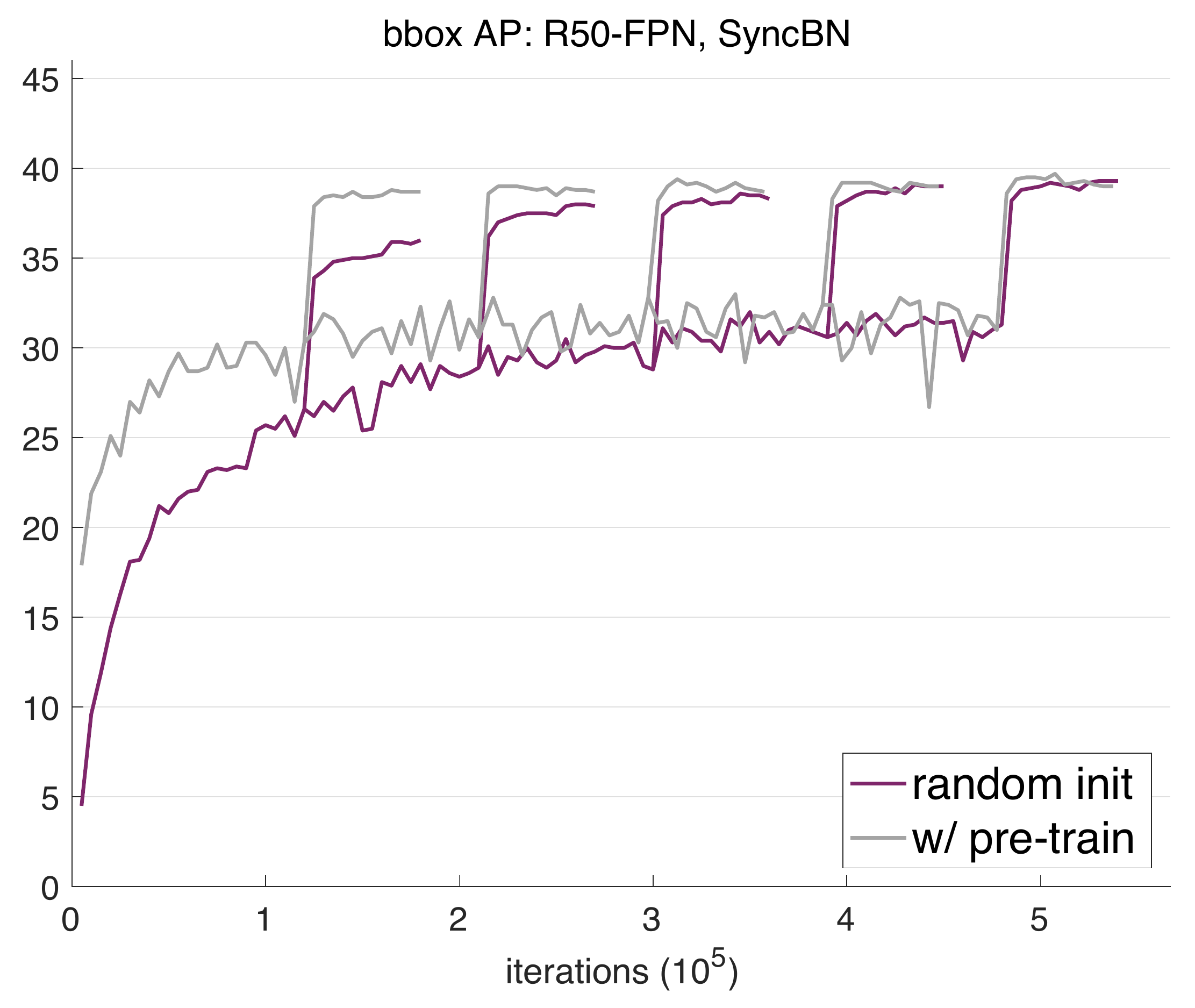}
\caption{Learning curves of AP$^\text{bbox}$ on COCO \texttt{val2017} using Mask R-CNN with R50-FPN and \textbf{SyncBN} \cite{Peng2018,Liu2018} (that synchronizes batch statistics across GPUs). The results of the 6$\times$ schedule are 39.3 (random initialization) and 39.0 (pre-training).
}\label{fig:bbox_r50_syncbn}
\end{figure}

\section{Results and Analysis}

\subsection{Training from scratch to match accuracy}

Our first surprising discovery is that when \emph{only} using the COCO data, models trained from scratch can catch up in accuracy with ones that are fine-tuned.

In this subsection, we train the models on the COCO \texttt{train2017} split that has $\app$118k (118,287) images, and evaluate in the 5k COCO \texttt{val2017} split. We evaluate bounding box (bbox) Average Precision (AP) for object detection and mask AP for instance segmentation.

\paragraph{Baselines with GN and SyncBN.} The validation bbox AP curves are shown in Figures~\ref{fig:bbox_r50_gn} and~\ref{fig:bbox_r101_gn} when using GN for ResNet-50 (R50) and ResNet-101 (R101) backbones and in Figure~\ref{fig:bbox_r50_syncbn} when using SyncBN for R50. For each figure, we compare the curves between models trained from random initialization \vs fine-tuned with ImageNet pre-training.

We study five different schedules for each case, namely, 2$\times$ to 6$\times$ iterations (Sec.~\ref{sec:settings}). Note that we overlay the five schedules of one model in the same plot. The leaps in the AP curves are a consequence of reducing learning rates, illustrating the results of different schedules.

Similar phenomena, summarized below, are consistently present in Figures~\ref{fig:bbox_r50_gn},~\ref{fig:bbox_r101_gn}, and~\ref{fig:bbox_r50_syncbn}:

\vspace{.5em}
\noindent\textbf{(i)} Typical fine-tuning schedules (2$\times$) work well for the models with pre-training to converge to near optimum (see also Table~\ref{tab:coco_gn_sched}, `w/ pre-train'). But these schedules are not enough for models trained from scratch, and they appear to be inferior if they are only trained for a short period.

\vspace{.5em}
\noindent\textbf{(ii)} Models trained from scratch \emph{can catch up} with their fine-tuning counterparts, if a 5$\times$ or 6$\times$ schedule is used---actually, when they converge to an optimum, their detection AP is \emph{no worse} than their fine-tuning counterparts.

\vspace{.5em}
In the standard COCO training set, ImageNet pre-training mainly helps to \emph{speed up convergence} on the target task early on in training, but shows \emph{little or no evidence} of improving the final detection accuracy. 

\renewcommand\arraystretch{1.1}
\setlength{\tabcolsep}{5pt}
\begin{table}[t]
\centering
\small
\begin{tabular}{cc|ccccc}
 \multicolumn{2}{c|}{schedule} & 2$\times$ & 3$\times$ & 4$\times$ & 5$\times$ & 6$\times$ \\
\shline
\multirow{2}{*}{R50} & random init & 36.8 & 39.5 & 40.6 & 40.7 & 41.3 \\
& w/ pre-train & 40.3 & 40.8 & 40.9 & 40.9 & 41.1 \\
\hline
\multirow{2}{*}{R101} & random init & 38.2 & 41.0 & 41.8 & 42.2 & 42.7 \\
& w/ pre-train & 41.8 & 42.3 & 42.3 & 41.9 & 42.2 \\
\end{tabular}
\vspace{.8em}
\caption{Object detection AP$^\text{bbox}$ on COCO \texttt{val2017} of training schedules from 2$\times$ (180k iterations) to 6$\times$ (540k iterations).
The model is Mask R-CNN with FPN and GN (Figures~\ref{fig:bbox_r50_gn} and~\ref{fig:bbox_r101_gn}).}
\label{tab:coco_gn_sched}
\end{table}

\renewcommand\arraystretch{1.1}
\setlength{\tabcolsep}{2.2pt}
\begin{table}[t]
\centering
\small
\begin{tabular}{cc|ccc|ccc}
& & AP$^\text{bbox}$ & AP$^\text{bbox}_\text{50}$ & AP$^\text{bbox}_\text{75}$ & AP$^\text{mask}$ & AP$^\text{mask}_\text{50}$ & AP$^\text{mask}_\text{75}$ \\
\shline
\multirow{3}{*}{R50} & random init & 41.3 & 61.8 & 45.6 & 36.6 & 59.0 & 38.9 \\
& w/ pre-train  & 41.1 & 61.7 & 44.6 & 36.4 & 58.5 & 38.7  \\
& $\triangle$ & \emph{+0.2} & \emph{+0.1} & \emph{+1.0} & \emph{+0.2} & \emph{+0.5} & \emph{+0.2} \\
\hline
\multirow{3}{*}{R101} & random init & 42.7 & 62.9 & 47.0 & 37.6 & 59.9 & 39.7 \\
& w/ pre-train & 42.3 & 62.6 & 46.2 & 37.2 & 59.7 & 39.7  \\
& $\triangle$ & \emph{+0.4} & \emph{+0.3} & \emph{+0.8} & \emph{+0.4} & \emph{+0.2} & \emph{0.0} \\
\end{tabular}
\vspace{.5em}
\caption{Training \textbf{from random initialization} \vs \textbf{with ImageNet pre-training} (Mask	R-CNN with FPN and GN, Figures~\ref{fig:bbox_r50_gn},~\ref{fig:bbox_r101_gn}), evaluated on COCO \texttt{val2017}. For each model, we show its results corresponding to the schedule (2 to 6$\times$) that gives the best AP$^\text{bbox}$.
}
\label{tab:coco_gn_aps}
\end{table}

\begin{figure}[t]
\centering
\includegraphics[width=.997\linewidth]{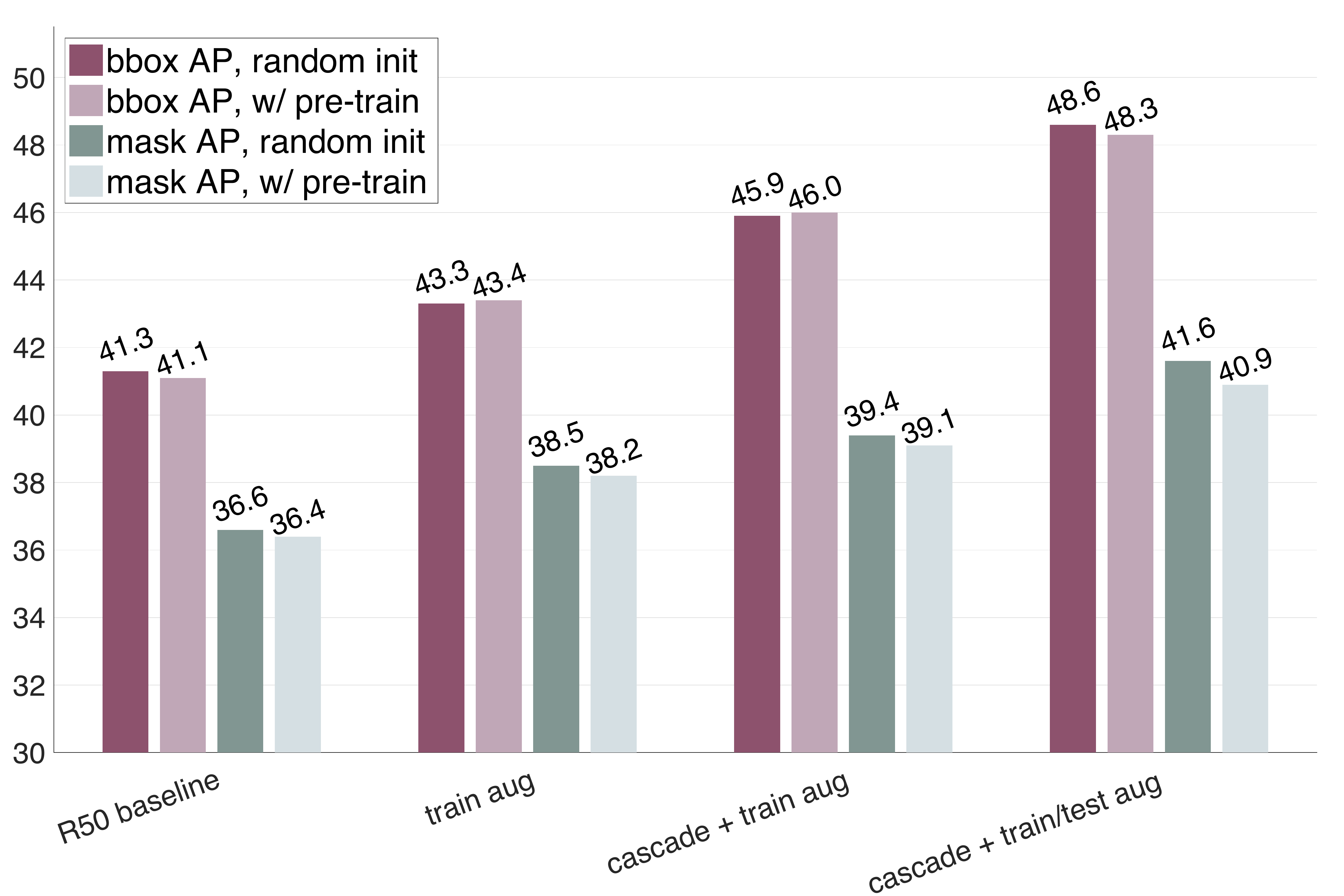}
\includegraphics[width=.997\linewidth]{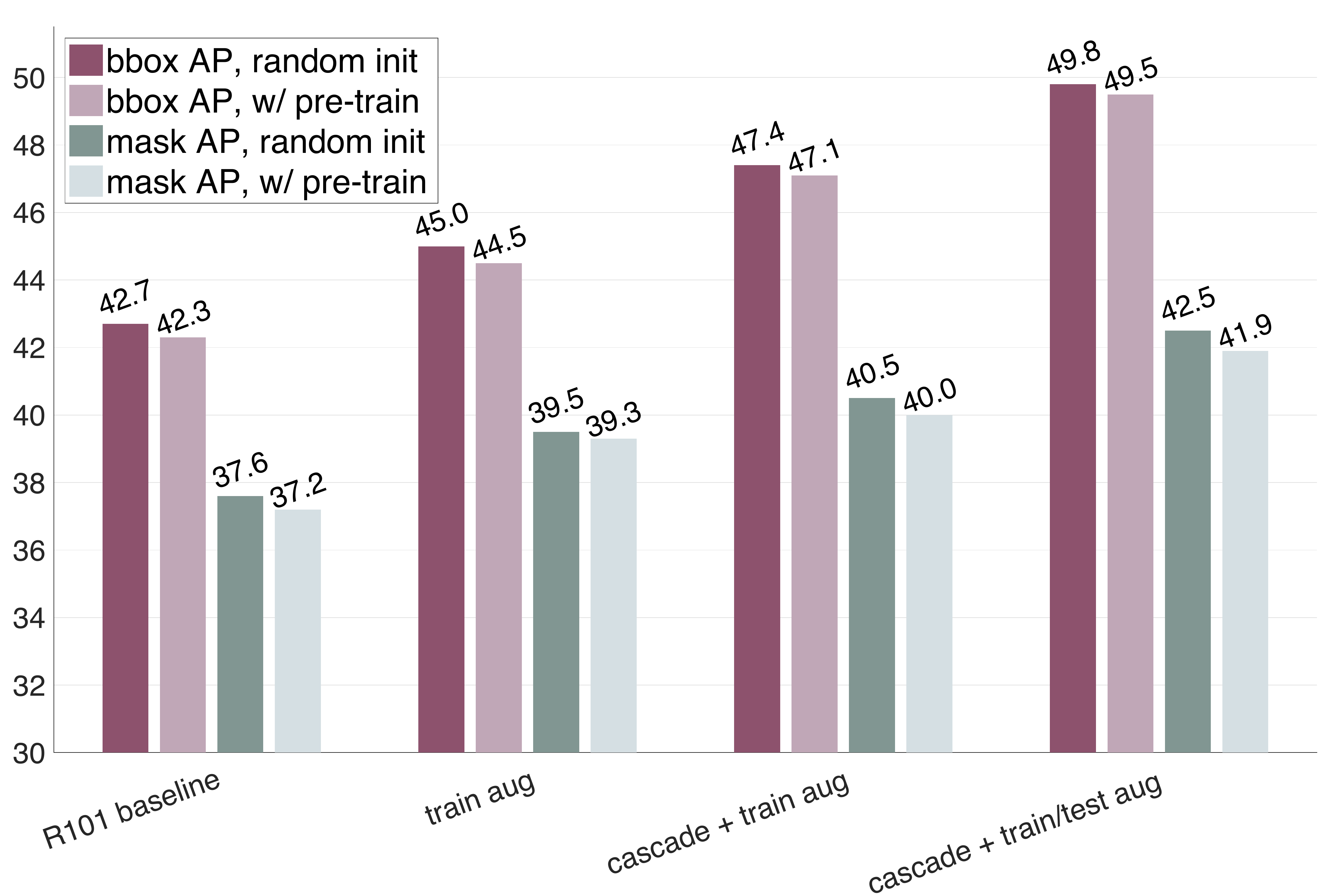}
\caption{Comparisons between \textbf{from random initialization \vs with pre-training} on various systems using Mask R-CNN, including: (i) baselines using FPN and GN, (ii) baselines with training-time multi-scale augmentation, (iii) baselines with Cascade R-CNN \cite{Cai2018} and training-time augmentation, and (iv) plus test-time multi-scale augmentation.
\textbf{Top: R50; Bottom: R101}. }\label{fig:allaround}
\vspace{-1em}
\end{figure}

\paragraph{Multiple detection metrics.} In Table~\ref{tab:coco_gn_aps} we further compare different detection metrics between models trained from scratch and with pre-training, including box-level and segmentation-level AP of Mask R-CNN, under Intersection-over-Union (IoU) thresholds of 0.5 (AP$_{50}$) and 0.75 (AP$_{75}$).

Table~\ref{tab:coco_gn_aps} reveals that models trained from scratch and with pre-training have \emph{similar} AP metrics under various criteria, suggesting that the models trained from scratch catch up not only by chance for a single metric.

Moreover, for the AP$^\text{bbox}_{75}$ metric (using a \emph{high overlap threshold}), training from scratch is better than fine-tuning by noticeable margins (1.0 or 0.8 AP).

\paragraph{Enhanced baselines.} The phenomenon that training with and without pre-training can be comparable is also observed in various enhanced baselines, as compared in Figure~\ref{fig:allaround}. We ablate the experiments as follows:

\newcommand{\bullets}[1]{\textbf{\emph{-- #1}}}

\vspace{.5em}
\bullets{Training-time scale augmentation}: Thus far all models are trained with no data augmentation except horizontal flipping. Next we use the simple training-time scale augmentation implemented in \Detectron: the shorter side of images is randomly sampled from [640, 800] pixels. Stronger data augmentation requires more iterations to converge, so we increase the schedule to 9$\times$ when training from scratch, and to 6$\times$ when from ImageNet pre-training.

Figure~\ref{fig:allaround} (`train aug') shows that in this case models trained with and without ImageNet pre-training are still comparable. Actually, stronger data augmentation relieves the problem of insufficient data, so we may expect that models with pre-training have less of an advantage in this case.

\vspace{.5em}
\bullets{Cascade R-CNN} \cite{Cai2018}: as a method  focusing on improving \emph{localization} accuracy, Cascade R-CNN appends two extra stages to the standard two-stage Faster R-CNN system. We implement its Mask R-CNN version by simply adding a mask head to the last stage. To save running time for the from-scratch models, we train Mask R-CNN from scratch without cascade, and switch to cascade in the final 270k iterations, noting that this does not alter the fact that the final model uses no ImageNet pre-training. We train Cascade R-CNN under the scale augmentation setting.

Figure~\ref{fig:allaround} (`cascade + train aug') again shows that Cascade R-CNN models have similar AP numbers with and without ImageNet pre-training. 
Supervision about \emph{localization} is mainly provided by the target dataset and is not explicitly available from the classification-based ImageNet pre-training. Thus we do not expect ImageNet pre-training to provide additional benefits in this setting.

\vspace{.5em}
\bullets{Test-time augmentation}: thus far we have used no test-time augmentation.
Next we further perform test-time augmentation by combining the predictions from multiple scaling transformations, as implemented in \Detectron \cite{Detectron2018}.

Again, the models trained from scratch are \emph{no worse} than their pre-training counterparts. Actually, models trained from scratch are even slightly better in this case---for example, mask AP is 41.6 (from scratch) \vs 40.9 for R50, and 42.5 \vs 41.9 for R101.

\paragraph{Large models trained from scratch.} We have also trained a significantly larger Mask R-CNN model from scratch using a ResNeXt-152 8$\times$32d \cite{Xie2017} (in short `X152') backbone with GN. The results are in Table~\ref{tab:coco_x152_gn}. 

This backbone has $\app$4$\times$ more FLOPs than R101. 
Despite being substantially larger, this model shows no noticeable overfitting. It achieves good results of \textbf{50.9} bbox AP and \textbf{43.2} mask AP in \texttt{val2017} when \emph{trained from random initialization}. We submitted this model to COCO 2018 competition, and it has \textbf{51.3} bbox AP and \textbf{43.6} mask AP in the \texttt{test-challenge} set. Our bbox AP is at the level of the COCO 2017 winners (50.5 bbox AP, \cite{Peng2018}), and is by far the highest number of its kind (single model, \emph{without ImageNet pre-training}).

We have trained the same model with ImageNet pre-training. It has bbox/mask AP of 50.3/42.5 in \texttt{val2017} (\vs from-scratch's 50.9/43.2). Interestingly, even for this large model, pre-training does not improve results.

\paragraph{\vs previous from-scratch results.}
DSOD \cite{Shen2017} reported 29.3 bbox AP by using an architecture specially tailored for results of training from scratch. A recent work of CornerNet \cite{Law2018} reported 42.1 bbox AP (w/ multi-scale augmentation) using no ImageNet pre-training. Our results, of various versions, are higher than previous ones. Again, we emphasize that previous works \cite{Shen2017,Law2018} reported \emph{no evidence} that models without ImageNet pre-training can be \emph{comparably} good as their ImageNet pre-training \emph{counterparts}.

\definecolor{demphcolor}{RGB}{144,144,144}
\newcommand{\demph}[1]{\textcolor{demphcolor}{#1}}
\renewcommand\arraystretch{1.1}
\setlength{\tabcolsep}{1.pt}
\begin{table}[t]
\centering
\small
\begin{tabular}{c|ccc|ccc}
& AP$^\text{bbox}$ & AP$^\text{bbox}_\text{50}$ & AP$^\text{bbox}_\text{75}$ & AP$^\text{mask}$ & AP$^\text{mask}_\text{50}$ & AP$^\text{mask}_\text{75}$ \\
\shline
\demph{R101 w/ train aug} & \demph{45.0} & \demph{65.7} & \demph{49.3} & \demph{39.5} & \demph{62.5} & \demph{42.1} \\
X152 w/ train aug & 46.4 & 67.1 & 51.1 & 40.5 & 63.9 & 43.4 \\
\quad+ cascade & 48.6 & 66.8 & 52.9 & 41.4 & 64.2 & 44.6 \\
\quad+ test aug & 50.9 & 68.7 & 55.4 & 43.2 & 66.1 & 46.8 \\
\end{tabular}
\vspace{.5em}
\caption{Mask R-CNN with \textbf{ResNeXt-152 trained from random initialization} (w/ FPN and GN), evaluated on COCO \texttt{val2017}.
}
\label{tab:coco_x152_gn}
\vspace{-.5em}
\end{table}

\begin{figure}[t]
\adjustbox{valign=t}{%
\begin{minipage}[c]{0.45\linewidth}
\centering
\includegraphics[width=1.\linewidth]{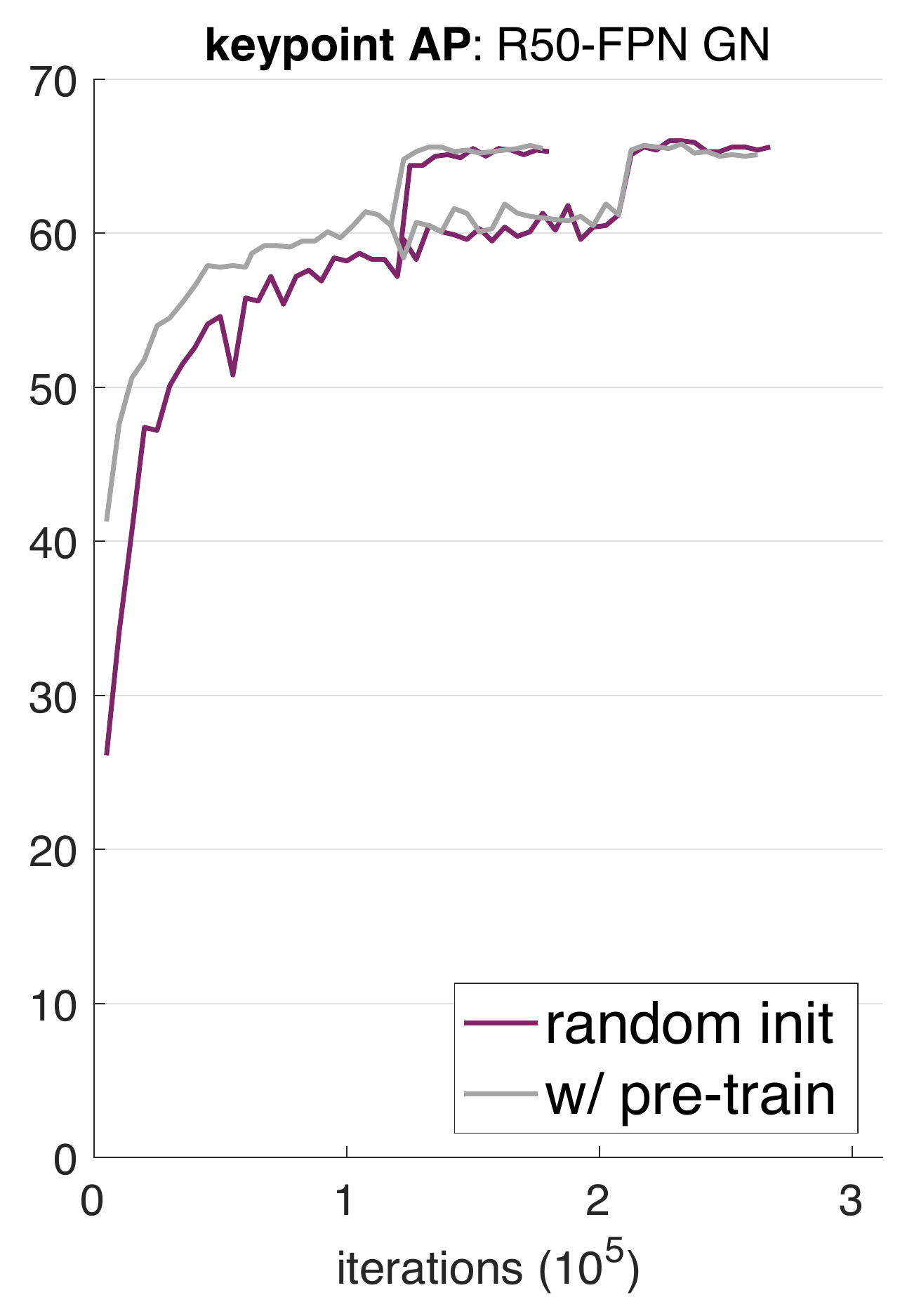}
\end{minipage}\hfill
}
\adjustbox{valign=t}{%
\begin{minipage}[c]{0.54\linewidth}
\vspace{.5em}
\caption{\textbf{Keypoint detection} on COCO using Mask R-CNN with R50-FPN and GN\@. We show keypoint AP on COCO \texttt{val2017}.
ImageNet pre-training has little benefit, and training from random initialization can quickly catch up \emph{without} increasing training iterations. We only need to use 2$\times$ and 3$\times$ schedules, unlike the object detection case.
The result is 65.6 \vs 65.5 (random initialization \vs pre-training) with 2$\times$ schedules.
}
\label{fig:bbox_r50_gn_kp}
\end{minipage}
}
\vspace{-1.5em}
\end{figure}

\paragraph{Keypoint detection.}
We also train Mask R-CNN for the COCO human keypoint detection task. The results are in Figure~\ref{fig:bbox_r50_gn_kp}. In this case, the model trained from scratch can catch up more quickly, and 
even when \emph{not} increasing training iterations, it is comparable with its counterpart that uses ImageNet pre-training.
Keypoint detection is a task more sensitive to fine spatial localization. Our experiment suggests that ImageNet pre-training, which has little explicit localization information, does not help keypoint detection.

\begin{figure*}[t]
\centering
\resizebox{.999\linewidth}{!}{
\centering
\includegraphics[height=19.8em]{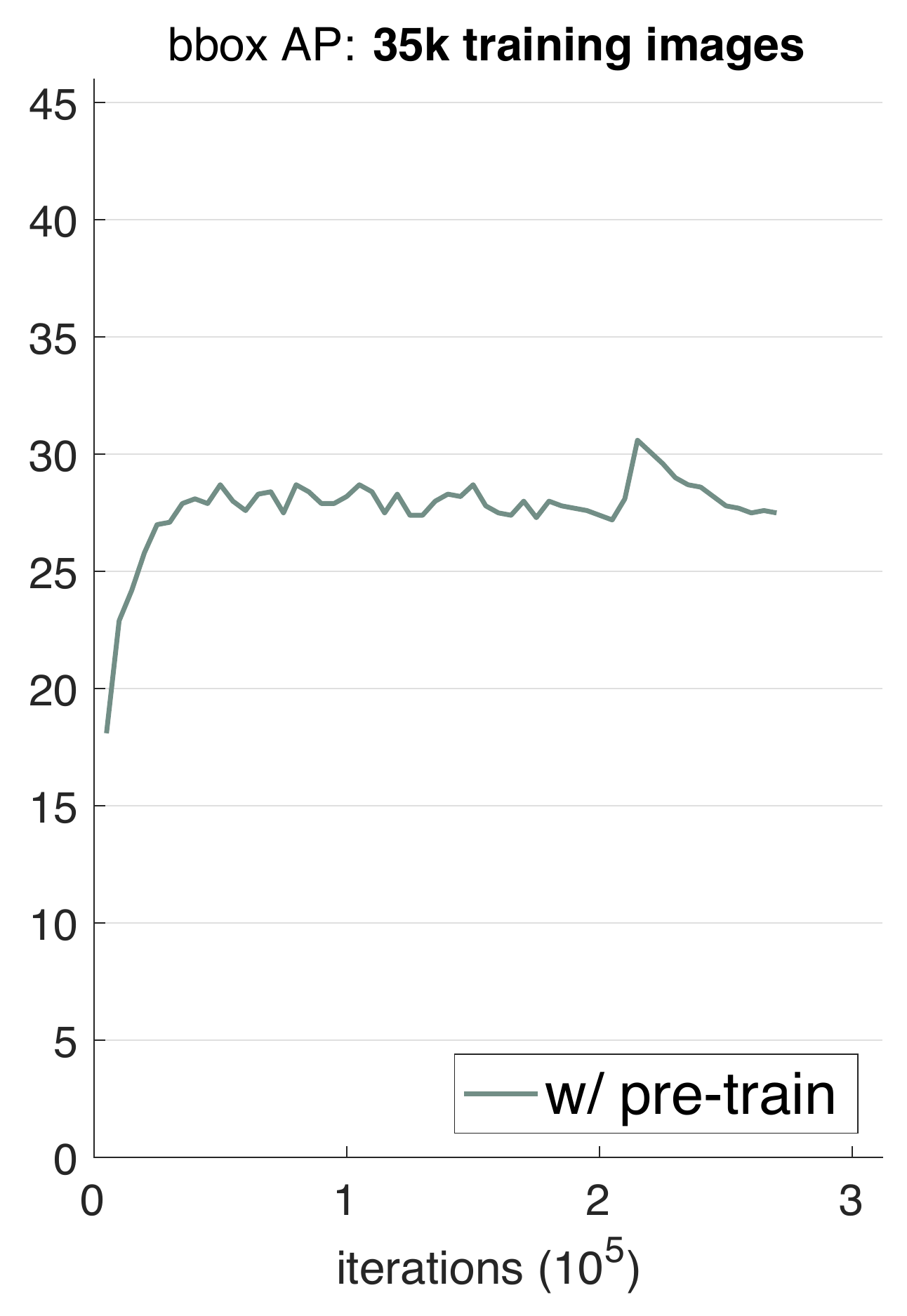}
\includegraphics[height=19.8em]{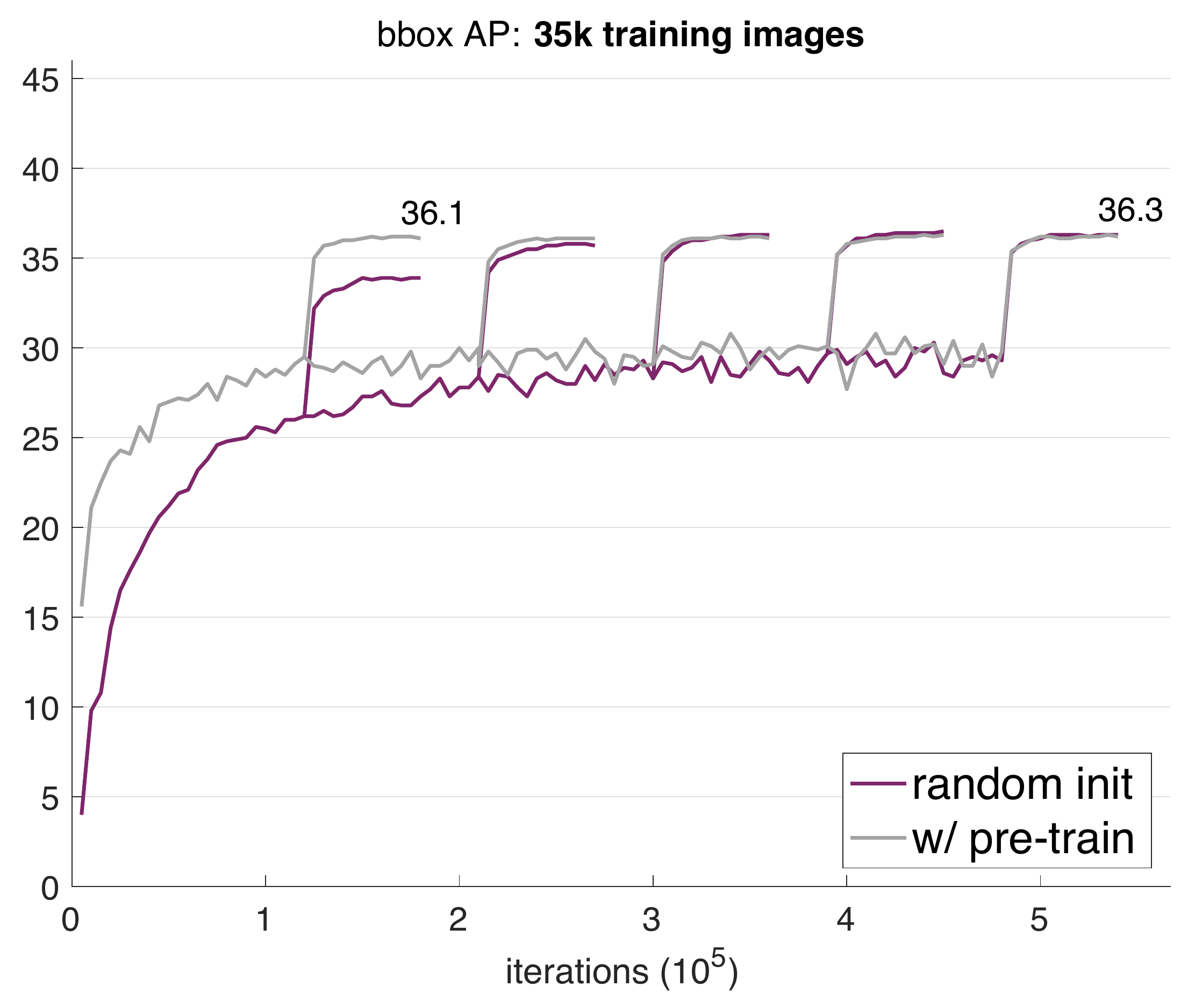}
\includegraphics[height=19.8em]{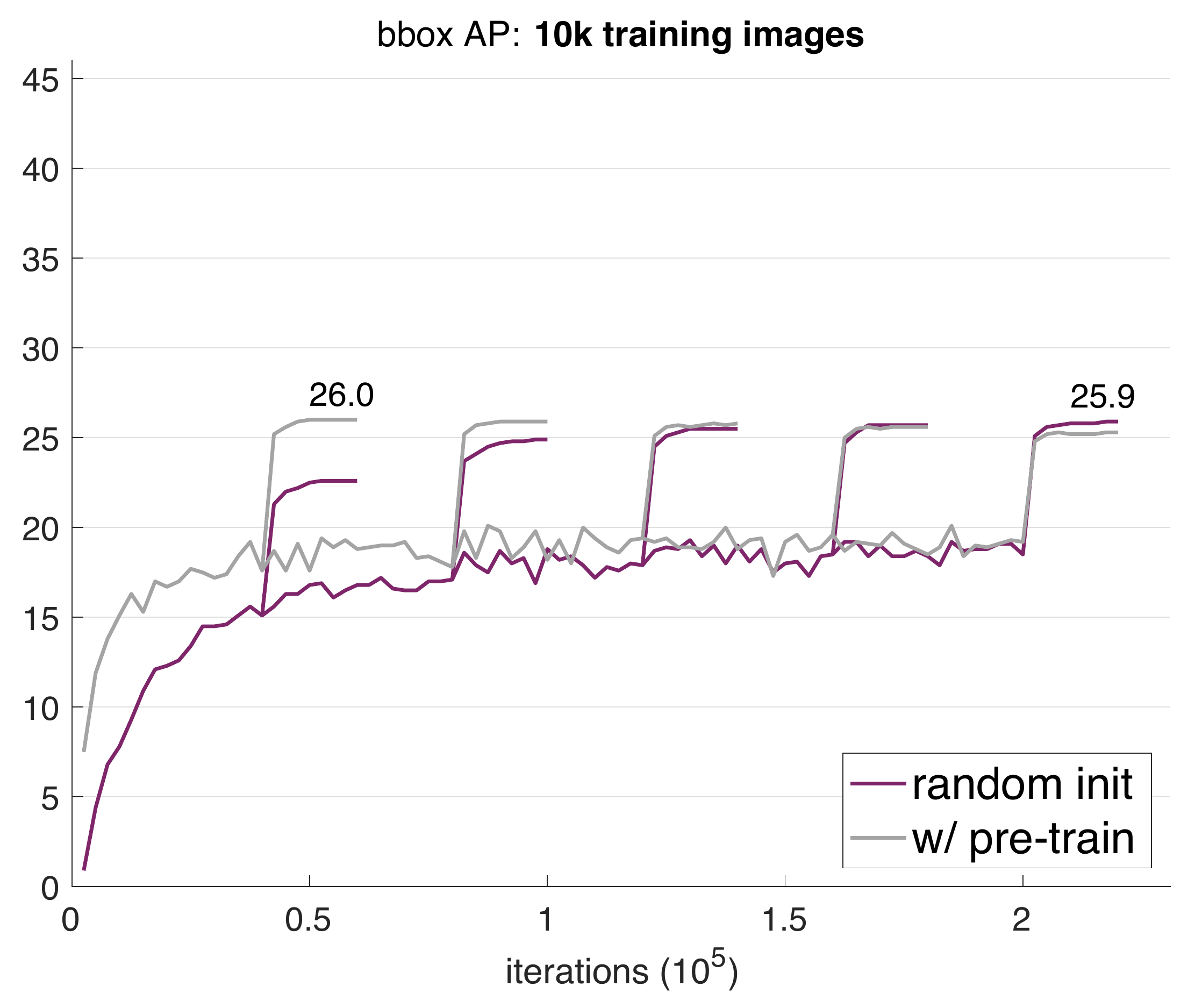}
}
\caption{\textbf{Training with fewer COCO images} (left/middle: \textbf{35k}; right: \textbf{10k}). 
The model is Mask R-CNN with R50-FPN and GN, evaluated by bbox AP in \texttt{val2017}.
\textbf{Left}: training with 35k COCO images, using the default hyper-parameters that were chosen for the 118k \texttt{train2017}. It shows overfitting before and after the learning rate changes.
\textbf{Middle}: training with 35k COCO images, using hyper-parameters optimized for `w/ pre-train' (the same hyper-parameters are then applied to the model from random initialization).
\textbf{Right}: training with 10k COCO images, using hyper-parameters optimized for `w/ pre-training'.
}\label{fig:bbox_r50_gn_subsets}
\end{figure*}

\paragraph{Models without BN/GN --- VGG nets.}
Thus far all of our experiments involve ResNet-based models, which require some form of activation normalization (\eg, BN or GN). Shallower models like VGG-16~\cite{Simonyan2015} can be trained from scratch without activation normalization as long as a proper initialization normalization is used~\cite{He2015}. Our next experiment tests the generality of our observations by exploring the behavior of training Faster R-CNN from scratch using VGG-16 as the backbone.

We implement the model following the original Faster R-CNN paper \cite{Ren2017} and its VGG-16 architecture; no FPN is used. 
We adopt standard hyper-parameters with a learning rate of 0.02, learning rate decay factor of 0.1, and weight decay of 0.0001. We use scale augmentation during training. Following previous experiments, we use the exact same hyper-parameters when fine-tuning and training from scratch. When randomly initializing the model, we use the same MSRA initialization~\cite{He2015} for ImageNet pre-training and for COCO from scratch.

The baseline model \emph{with pre-training} is able to reach a maximal bbox AP of 35.6 after an extremely long 9$\times$ training schedule (training for longer leads to a slight degradation in AP\@). Here we note that even \emph{with} pre-training, full convergence for VGG-16 is slow. The model trained from scratch reaches a similar level of performance with a maximal bbox AP of 35.2 after an 11$\times$ schedule (training for longer resulted in a lower AP, too). These results indicate that our methodology of `making minimal/no changes' (Sec.~\ref{sec:method}) but adopting good optimization strategies and training for longer are sufficient for training comparably performant detectors on COCO, compared to the standard `pre-training and fine-tuning' paradigm.

\subsection{Training from scratch with less data}

Our second discovery, which is even more surprising, is that with substantially less data (\eg, $\app$1/10 of COCO), models trained from scratch are \emph{no worse} than their counterparts that are pre-trained.

\paragraph{35k COCO training images.} We start our next investigation with $\app$1/3 of COCO training data (35k images from \texttt{train2017}, equivalent to the older \texttt{val35k}). We train models with or without ImageNet pre-training on this set.

Figure~\ref{fig:bbox_r50_gn_subsets} (left) is the result using ImageNet pre-training under the hyper-parameters of Mask R-CNN that were chosen for the 118k COCO set. These hyper-parameters are not optimal, and the model suffers from overfitting even with ImageNet pre-training. It suggests that \emph{ImageNet pre-training does not automatically help reduce overfitting}.

To obtain a healthy baseline, we redo grid search for hyper-parameters on the models that are \emph{with} ImageNet pre-training.\footnote{Our new recipe changes are: training-time scale augmentation range of [512, 800] (\vs baseline's no scale augmentation), a starting learning rate of 0.04 (\vs 0.02), and a learning rate decay factor of 0.02 (\vs 0.1).} The gray curve in Figure~\ref{fig:bbox_r50_gn_subsets} (middle) shows the results. It has optimally 36.3 AP with a 6$\times$ schedule.

Then we train our model from scratch using \emph{the exact same} new hyper-parameters that are chosen for the pre-training case.
This obviously biases results in favor of the pre-training model.
Nevertheless, the model trained from scratch has 36.3 AP and \emph{catches up} with its pre-training counterpart (Figure~\ref{fig:bbox_r50_gn_subsets}, middle), despite less data.

\paragraph{10k COCO training images.} We repeat the same set of experiments on a smaller training set of 10k COCO images (\ie, \emph{less than 1/10th of the full COCO set}). Again, we perform grid search for hyper-parameters on the models that use ImageNet pre-training, and apply them to the models trained from scratch. We shorten the training schedules in this small training set (noted by x-axis, Figure~\ref{fig:bbox_r50_gn_subsets}, right).

The model with pre-training reaches 26.0 AP with 60k iterations, but has a slight degradation when training more. The counterpart model trained from scratch has 25.9 AP at 220k iterations, which is \emph{comparably} accurate.

\paragraph{Breakdown regime: 1k COCO training images.} That training from scratch in 10k images is comparably accurate is surprising. But it is not reasonable to expect this trend will last for arbitrarily small target data, as we report next.

\begin{figure}[t]
\centering
\includegraphics[width=.9\linewidth]{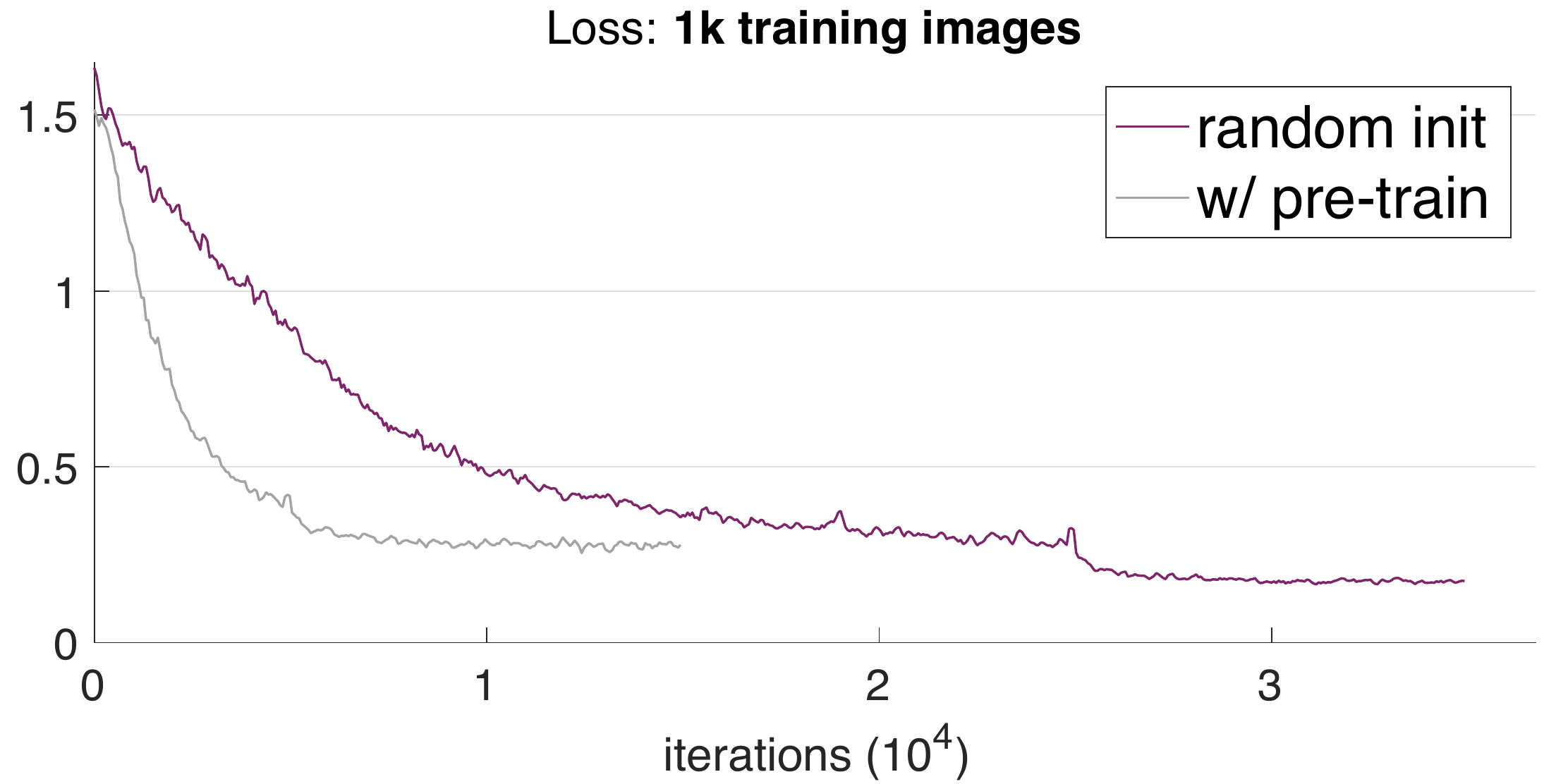}
\caption{Training with \textbf{1k COCO images} (shown as the \emph{loss} in the training set). The model is Mask R-CNN with R50-FPN and GN\@.
As before, we use hyper-parameters optimized for the model with pre-training, and apply the same hyper-parameters to the model from random initialization. The randomly initialized model can catch up for the \emph{training loss}, but has lower \emph{validation accuracy} (3.4 AP) than the pre-training counterpart (9.9 AP).
}\label{fig:loss_r50_gn_val1k}
\vspace{-.8em}
\end{figure}

In Figure~\ref{fig:loss_r50_gn_val1k} we repeat the same set of experiments using only 1k COCO training images ($\app$1/100th of full COCO, again optimizing hyper-parameters for the pre-training case) and show the training loss. In terms of \emph{optimization} (\ie, reducing training loss), training from scratch is still \emph{no worse} but only converges more slowly, as seen previously. However, in this case, the training loss does not translate into a good validation AP: the model with ImageNet pre-training has 9.9 AP \vs the from scratch model's 3.5 AP\@.
For one experiment only we also performed a grid search to optimize the from-scratch case: the result improves to 5.4 AP, but does not catch up\@. This is a sign of strong overfitting due to the severe lack of data.

We also do similar experiments using 3.5k COCO training images. The model that uses pre-training has a peak of 16.0 bbox AP \vs the trained from scratch counterpart's 9.3 AP. The breakdown point in the COCO dataset is somewhere between 3.5k to 10k training images.

\paragraph{Breakdown regime: PASCAL VOC.} Lastly we report the comparison in PASCAL VOC object detection \cite{Everingham2010}. We train on the set of \texttt{trainval2007}+\texttt{train2012}, and evaluate on \texttt{val2012}.
Using ImageNet pre-training, our Faster R-CNN baseline (with R101-FPN, GN, and only training-time augmentation) has 82.7 mAP at 18k iterations. Its counterpart trained from scratch in VOC has 77.6 mAP at 144k iterations and does not catch up even training longer.

There are 15k VOC images used for training. But these images have on average 2.3 instances per image (\vs COCO's $\app$7) and 20 categories (\vs COCO's 80). They are not directly comparable to the same number of COCO images. We suspect that the fewer instances (and categories) has a similar negative impact as insufficient training data, which can explain why training from scratch on VOC is not able to catch up as observed on COCO.

\section{Discussions}

We summarize the main observations from our experiments as follows:

- \emph{Training from scratch on target tasks is possible without architectural changes.}

- \emph{Training from scratch requires more iterations to sufficiently converge.}

- \emph{Training from scratch can be no worse than its ImageNet pre-training counterparts under many circumstances, down to as few as 10k COCO images}.

- \emph{ImageNet pre-training speeds up convergence on the target task}.

- \emph{ImageNet pre-training does not necessarily help reduce overfitting unless we enter a very small data regime.}

- \emph{ImageNet pre-training helps less if the target task is more sensitive to localization than classification.}

\vspace{1em}
Based on these observations, we provide our answers to a few important questions that may encourage people to rethink ImageNet pre-training:

\emph{\textbf{Is ImageNet pre-training necessary?}} No---if we have enough target data (and computation). Our experiments show that ImageNet can help speed up convergence, but does not necessarily improve accuracy unless the target dataset is too small (\eg, $<$10k COCO images). It can be \emph{sufficient} to directly train on the target data if its dataset scale is large enough. Looking forward, this suggests that \emph{collecting annotations of target data (instead of pre-training data)} can be more useful for improving the target task performance.

\emph{\textbf{Is ImageNet helpful?}} Yes. ImageNet pre-training has been a \emph{critical} auxiliary task for the computer vision community to progress. 
It enabled people to see significant improvements before larger-scale data was available (\eg, in VOC for a long while).
It also largely helped to circumvent optimization problems in the target data (\eg, under the lack of normalization/initialization methods). Moreover, ImageNet pre-training reduces research cycles, leading to \emph{easier} access to encouraging results---pre-trained models are widely and freely available today, pre-training cost does not need to be paid repeatedly, and fine-tuning from pre-trained weights converges faster than from scratch.
We believe that these advantages will still make ImageNet undoubtedly helpful for computer vision research.

\emph{\textbf{Do we need big data?}} Yes. But a generic large-scale, \emph{classification-level} pre-training set is not ideal if we take into account the extra effort of collecting and cleaning data---the cost of collecting ImageNet has been largely ignored, but the `pre-training' step in the `pre-training + fine-tuning' paradigm is in fact \emph{not free} when we scale out this paradigm. If the gain of large-scale classification-level pre-training becomes exponentially diminishing \cite{Sun2017,Mahajan2018}, it would be more effective to collect data in the target domain.

\emph{\textbf{Shall we pursuit universal representations?}} Yes. We believe learning universal representations is a laudable goal. Our results do not mean deviating from this goal. Actually, our study suggests that the community should be more careful when evaluating pre-trained features (\eg, for \emph{self-supervised} learning \cite{Doersch2015,Wang2015a,Pathak2016,Pathak2017}), as now we learn that even random initialization could produce excellent results.

\vspace{.5em}

In closing, ImageNet and its pre-training role have been incredibly influential in computer vision, and we hope that our new experimental evidence about ImageNet and its role will shed light into potential future directions for the community to move forward.

{\small
\bibliographystyle{ieee}
\bibliography{scratch}
}

\end{document}